\documentclass[10pt]{wlscirep}
\usepackage[utf8]{inputenc}
\usepackage[T1]{fontenc}
\usepackage{makecell} % 放在导言区
\usepackage{multirow}

\title{Dissecting Role Cognition in Medical LLMs via Neuronal Ablation}
\usepackage{CJKutf8}
\author[1,+]{Xun Liang}
\author[1,+]{Huayi Lai}
\author[1]{Hanyu Wang}
\author[2]{Wentao Zhang}
\author[3]{Linfeng Zhang}
\author[4,*]{Yanfang Chen}
\author[5,6]{Feiyu Xiong}
\author[5,6,*]{Zhiyu Li}
\affil[1]{School of Information, Renmin University of China, Beijing, China}
\affil[2]{Center of Machine Learning Research, Peking University, Beijing, China}
\affil[3]{School of Artificial Intelligence, Shanghai Jiaotong University, Shanghai, China}
\affil[4]{Health Informatics Committee of the China Society for Scientific and Technical Information, Renmin University of China, Beijing, China}
\affil[5]{Institute for Advanced Algorithms Research (IAAR), Shanghai, China}
\affil[6]{Memtensor Research Center, Shanghai, China}
\affil[*]{cyf@ruc.edu.cn,lizy@iaar.ac.cn}
\affil[+]{They contribute equally to this work}

\begin{abstract}
Large language models (LLMs) have gained significant traction in medical decision support systems, particularly in the context of medical question answering and role-playing simulations. A common practice, Prompt-Based Role Playing (PBRP), instructs models to adopt different clinical roles (e.g., medical students, residents, attending physicians) to simulate varied professional behaviors. However, the impact of such role prompts on model reasoning capabilities remains unclear. This study introduces the  RP-Neuron-Activated Evaluation Framework(RPNA) to evaluate whether role prompts induce distinct, role-specific cognitive processes in LLMs or merely modify linguistic style. We test this framework on three medical QA datasets, employing neuron ablation and representation analysis techniques to assess changes in reasoning pathways. Our results demonstrate that role prompts do not significantly enhance the medical reasoning abilities of LLMs. Instead, they primarily affect surface-level linguistic features, with no evidence of distinct reasoning pathways or cognitive differentiation across clinical roles. Despite superficial stylistic changes, the core decision-making mechanisms of LLMs remain uniform across roles, indicating that current PBRP methods fail to replicate the cognitive complexity found in real-world medical practice. This highlights the limitations of role-playing in medical AI and emphasizes the need for models that simulate genuine cognitive processes rather than linguistic imitation.We have released the related code in the following repository:\url{https://github.com/IAAR-Shanghai/RolePlay_LLMDoctor}
\end{abstract}

\begin{document}
\flushbottom
\maketitle
\thispagestyle{empty}

\section{Introduction}
\begin{figure}
    \centering
    \includegraphics[width=1\linewidth]{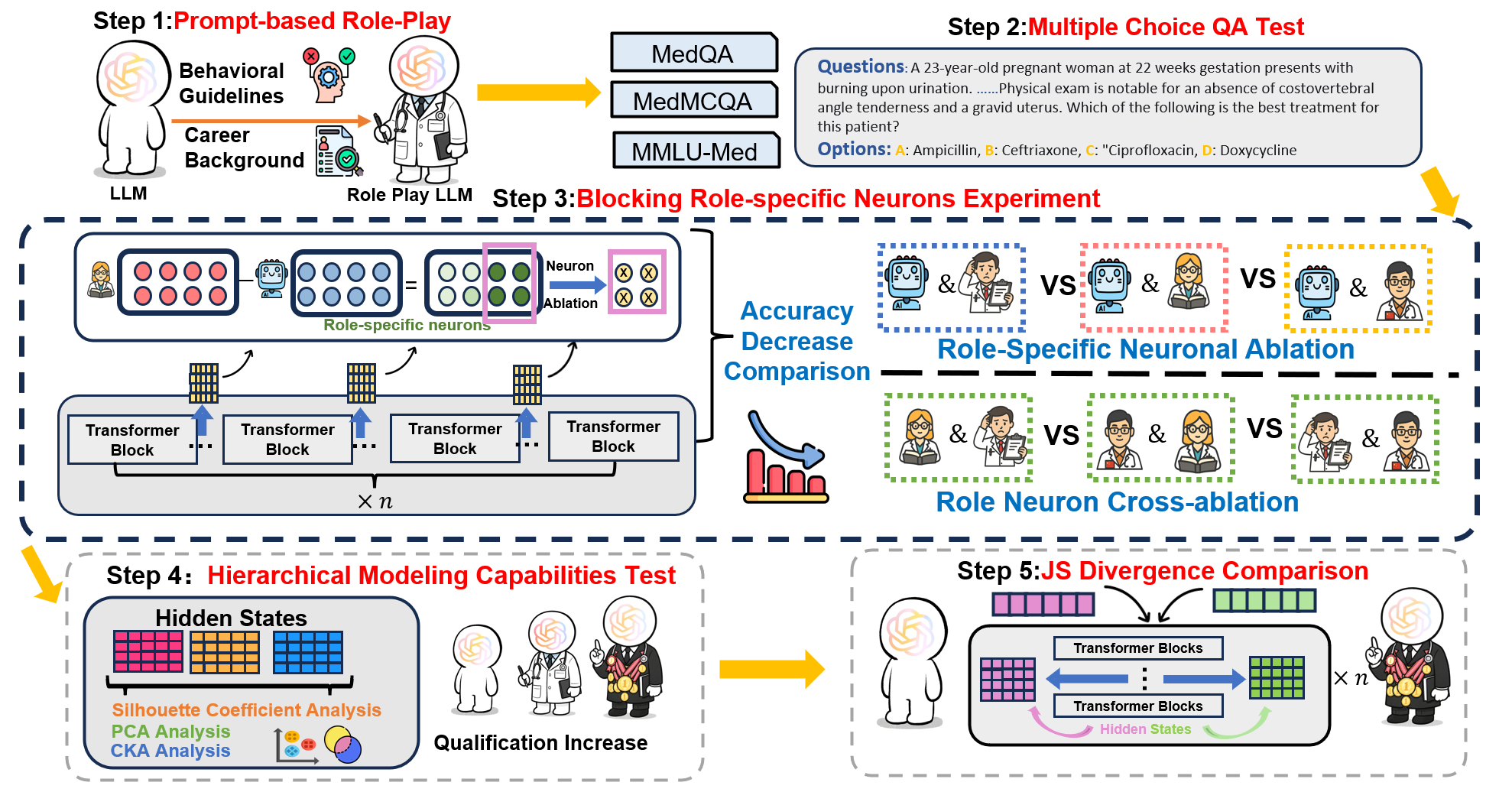}
    \caption{Overview of the RPNA evaluation framework.}
    \label{fig:pipeline}
\end{figure}
Driven by the rapid advancement of large language models(LLMs)\cite{sr1,sr2,sr3,sr4}, LLMs are increasingly used for medical question answering and clinical decision support\cite{MedicalFuture1,MedicalFuture2,MedicalFuture3,LLMsurvey}. Mainstream architectures include autoregressive generators such as the GPT series\cite{GPT4} and encoder–decoder models such as the T5 family\cite{T5,Transformer}. These models perform well on medical QA\cite{MedicalHealthQA}, case summarization\cite{MedicalSummary}, and patient-facing dialogue\cite{jMedicalConversation}. A prevalent practice in these applications is Prompt-Based Role Playing (PBRP)\cite{RP,LLMPromptRoles1}, which instructs a model to respond “as” a particular clinician (e.g., intern, resident, attending, or specialist) with the goal of increasing realism and credibility of generated outputs. Building on PBRP, recent systems adopt multi-agent designs where prompt-conditioned agents coordinate via tool use and deliberation to improve decisions\cite{agenthospital,metagpt}. Despite their appeal, the validity of these agent configurations (role prompts, division of labor, and interaction protocols) and their actual contribution to reasoning quality remain insufficiently tested. We introduce the RP-Neuron-Activated Evaluation Framework(RPNA) to test whether role prompts induce complementary, role-specific computation or merely stylistic modulation.

In clinical medicine, role stratification is more than an allocation of authority or labor; it encodes graded expertise with heterogeneous cognitive mechanisms, reasoning strategies, and knowledge structures across levels\cite{DoctorJobs,DoctorJobs2}. From novice interns to frontline attendings and senior specialists, clinicians deploy different diagnostic heuristics, evidence-integration routines, and risk-calibration policies when facing the same case. Empirical work in cognitive psychology and statistical decision theory has documented systematic differences in neural activation and information-processing policies across expertise levels during history taking, symptom interpretation, and the construction of diagnostic–therapeutic pathways\cite{DoctorJobs5,DoctorJobs4,DoctorJobs3,Doctorjobs6}. This stratified cognition underpins medical education’s role progression and enables multi-level, collaborative decision making in practice. Therefore, if LLMs “play” clinical roles, a testable expectation is role-dependent differences in latent computation and behavior, beyond surface style.

A central, under-tested question is whether prompt-based role playing actually induces role-specific reasoning in LLMs or merely produces stylistic shifts while leaving latent computations largely unchanged. We address this with the RP-Neuron-Activated Evaluation Framework(RPNA) (As shown in Figure \ref{fig:pipeline}). We identify role-salient units using an activation/gradient-based salience score under each role prompt, then ablate the top-r\% units in the top-K layers during inference. If role prompts instantiate distinct circuits, within-role masking should degrade performance more than cross-role masking; symmetry or uniformly small drops argues against role-specific pathways. Around this probe, RPRF benchmarks pre/post-prompt accuracy, assesses representation structure (CKA/PCA and clustering), and profiles layer-wise divergence via JSD across depth.

We apply RPNA on 3 medical QA datasets on 3 kinds of open-source LLMs with different model size. To disentangle style from computation, we fix decoding, compute paired accuracy deltas with 95\% confidence intervals, and sweep masking ratios. RPRF then asks four testable questions: Do role prompts yield consistent accuracy gains? Do role effects persist in deep layers rather than fading with depth? Do role conditions form separable clusters in representation space (CKA/PCA and clustering)? and is within-role neuron masking more damaging than cross-role masking? Failure on these indicates that role prompts mainly coordinate language style rather than reconfigure reasoning.

Our findings indicate that role prompts do not induce significant changes in latent reasoning pathways across clinical roles, and any observed accuracy improvements are unstable. Neuron masking for a specific role results in performance degradation comparable to that observed for other roles, and cross-role masking produces effects similar to within-role masking, providing no evidence for the existence of role-specific circuits. These results suggest substantial homogeneity in latent computation across clinical roles under PBRP, with prompts primarily influencing surface language rather than inducing distinct reasoning pathways. Activation differences are concentrated in early to middle layers and attenuate with depth; high CKA values and overlapping PCA clusters indicate convergent deep representations. Collectively, these findings highlight that role playing in PBRP mainly tunes linguistic style rather than contributing to genuine improvements in reasoning. Consequently, the Role-Prompted Neuron-Activated Evaluation Framework (RPNA) provides a clear and practical foundation for assessing when role-playing agent pipelines are likely to deliver true advancements in clinical decision support, as opposed to merely enhancing stylistic realism.

As shown in the pipeline in Figure \ref{fig:pipeline}, steps 1 to 5 outline the workflow of the RPNA framework. In Step 1, PBRP integrates LLMs with specific behavioral guidelines and career backgrounds to simulate reasoning processes across clinical roles. Step 2 evaluates the model’s performance through multiple-choice QA tests, posing questions and reasoning over different options. In Step 3, a lightweight, model-agnostic neuron-masking ablation method is applied to identify role-salient neural units and compare the effects of within-role versus cross-role masking, revealing no evidence of role-specific circuits. This suggests that role prompts mainly tune linguistic style rather than alter reasoning paths. In Step 4, representation structures are examined using CKA/PCA analysis and clustering, revealing convergent deep representations across roles rather than forming separable, hierarchy-aligned groups. Finally, Step 5 profiles depth-wise divergence using layer-wise Jensen-Shannon Divergence (JSD), showing that any role effects are concentrated in early to middle layers and attenuate with depth. These results provide actionable guidance for determining when single- or multi-agent role-playing pipelines offer genuine improvements in clinical decision support, as opposed to merely enhancing stylistic realism.

\section*{Methods}

\begin{figure}
    \centering
    \includegraphics[width=1\linewidth]{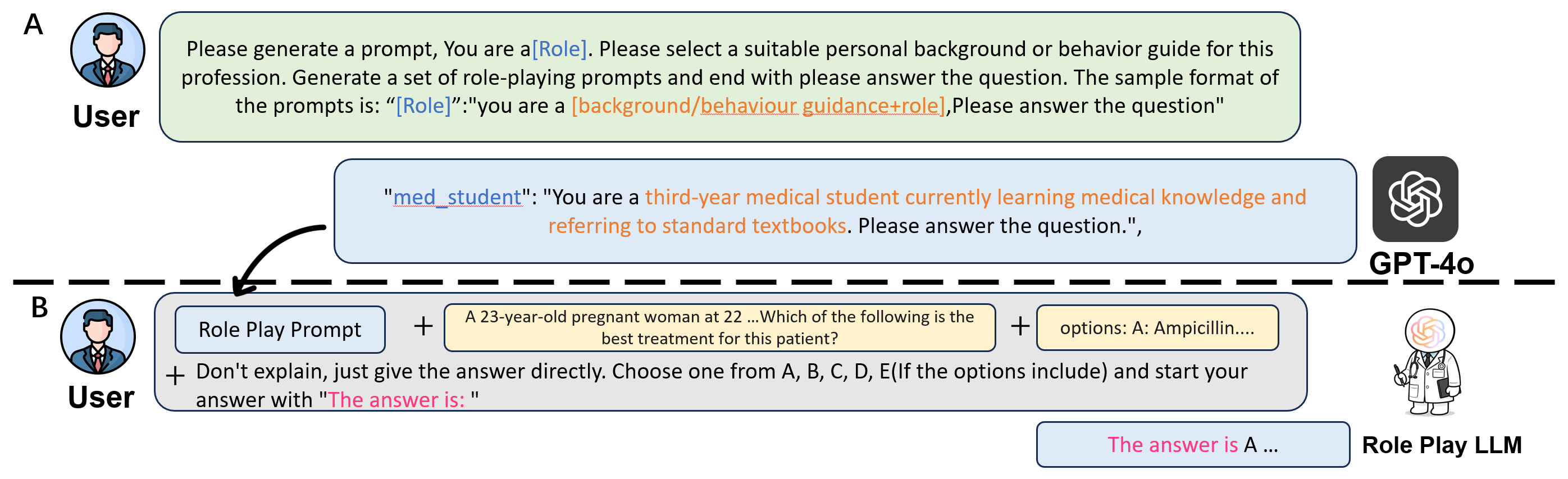}
    \caption{ The figure shows the
 overall process of constructing role-playing prompts (Role Prompts) in the Q\&A task.Figure A shows the step of creating a standard prompt for a Medical Student ,Figure B simulates the reasoning behavior of real medical professionals under different knowledge backgrounds and thinking styles. }
    \label{fig:role_prompt}
\end{figure}
\subsection{Dataset Selection}
\begin{figure}
    \centering
    \includegraphics[width=1\linewidth]{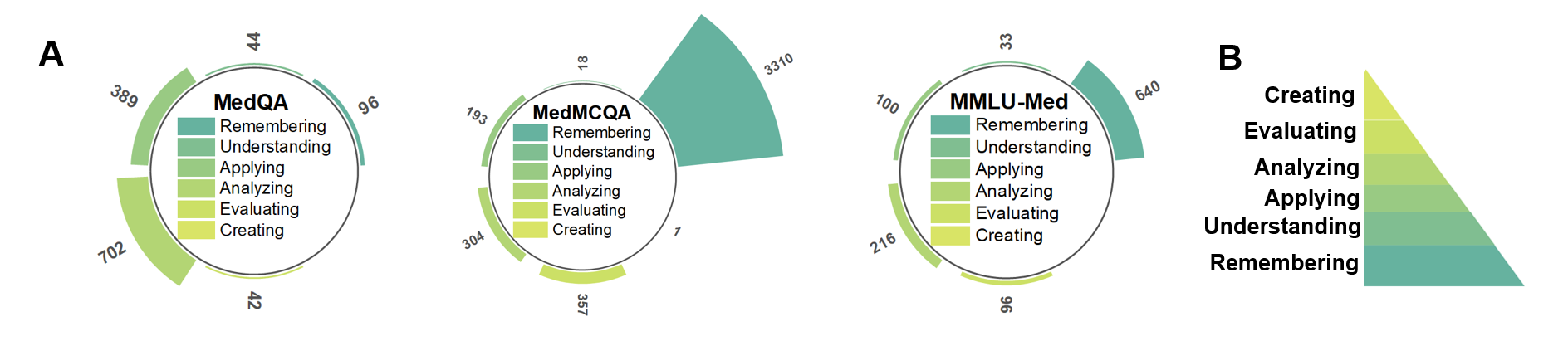}
    \caption{Task classification of three medical question-answering datasets (based on GPT-4o). Figure A shows the classification results based on Bloom's taxonomy, and Figure B shows the six levels of Bloom's taxonomy from high to low.}
    \label{fig:dataset}
\end{figure}

To comprehensively assess the cognitive modeling capabilities of large language models (LLMs) in medical role-playing, we selected three representative and complementary medical question-answering datasets, covering basic knowledge, clinical reasoning, and interdisciplinary scenarios. This selection ensures a balanced assessment of tasks in terms of hierarchy and cognitive complexity, as categorized using Bloom's Taxonomy (As shown in Figure \ref{fig:dataset}.B). The datasets span a range of cognitive levels, from fundamental recall-based tasks to more complex clinical reasoning and interdisciplinary integration (As shown in Figure \ref{fig:dataset}.A).

\begin{itemize}
\item \textbf{MedQA}\cite{medqa}: This dataset is derived from examination papers of the medical boards in the United States, mainland China, and Taiwan. It is designed to assess doctors' professional knowledge and clinical decision-making abilities. The dataset contains numerous real-world clinical questions across fields such as internal medicine, surgery, pediatrics, gynecology, and radiology. The question design emphasizes multi-step reasoning and knowledge integration, making it crucial for evaluating the model's higher-order reasoning capabilities. From MedQA, we selected the USMLE sub-dataset for testing, containing 1,273 multiple-choice questions in the test set. As shown in Figure \ref{fig:dataset}, the majority of the questions in MedQA are categorized under "Remembering" and "Understanding" in Bloom's Taxonomy, with a smaller proportion requiring higher cognitive levels such as "Evaluating" and "Creating".

\item \textbf{MedMCQA}\cite{medmcqa}: Derived from the Indian NEET and AIIMS medical exam question banks, this dataset focuses on basic medical courses such as anatomy, physiology, and biochemistry. The questions are structured in a clear, standard format, making it suitable for testing the model's ability to understand medical terminology and textbook-level knowledge. We used the test set of this dataset, which contains 4,183 multiple-choice questions. As depicted in Figure \ref{fig:dataset}, MedMCQA questions predominantly fall under "Remembering" and "Understanding", with only a few questions in the "Applying" and "Analyzing" categories, reflecting the focus on foundational knowledge.

\item \textbf{MMLU-Med}\cite{mmlu}: Focused on interdisciplinary knowledge integration, MMLU-Med tests models on common-sense fusion and conceptual reasoning abilities. We selected the medical-related sub-datasets from MMLU, covering topics like anatomy, clinical knowledge, professional medicine, medical genetics, college medicine, and college biology, totaling 1,083 multiple-choice questions. These sub-datasets include a wide range of cognitive levels, with a significant proportion of questions in the higher-order categories of "Analyzing", "Evaluating", and "Creating", as shown in Figure \ref{fig:dataset}, aligning with the need for more complex reasoning in interdisciplinary medical contexts.

\end{itemize}

MedQA and MMLU-Med use the standard four-option multiple-choice question format, while MedMCQA utilizes a five-option format. We selected the test set portion from each dataset to ensure the performance results are as accurate and consistent as possible. The distribution of questions across different levels of Bloom's Taxonomy ensures that the datasets provide a comprehensive challenge for assessing LLMs' cognitive reasoning in medical contexts.
\subsection{Prompt Configuration}

To systematically evaluate the impact of different doctor role settings on the behavior and internal representations of LLMs, as shown in Figure\ref{fig:role_prompt}, we constructed a prompt set containing multiple role contexts, divided into three major categories: Role-Playing Group, Baseline Group, and Control Group. The QA example of using the role playing prompt can be seen in the code repository.

\paragraph{Role-Playing Group (Role-Playing Prompts)}
Role-playing prompts simulate the differences in knowledge, experience, and decision-making styles across medical professionals. These prompts are designed to test if the model can reflect role-specific reasoning. We created ten representative doctor roles based on medical education and clinical role division standards, generating the prompts using GPT\-4o (Figure \ref{fig:role_prompt}.A). The prompts include “[Background/Behavioral Guidance + Role Name] + Please answer the question,” guiding the model to answer as a specific role. Figure \ref{fig:role_prompt}.B shows how the role prompt is combined with the clinical Q\&A task and output constraints to guide the model's reasoning. 

\paragraph{Baseline Group (Baseline Prompts)}
To establish a baseline, we used a unified prompt: “Please provide the most appropriate answer to the following medical question.” This prompt lacks role-specific information, allowing the model to respond based on its default knowledge.

\paragraph{Control Group (Random Prompts)}
To test the model's sensitivity to medical contexts, we included random, non-medical prompts (e.g., \textbf{“This is a sentence.”}). These prompts are not related to the task and are expected to have no impact on the model’s answers, serving as a measure of the model’s robustness and the selectivity of the induction mechanism.

\subsection{Model Selection and Setup}  
To systematically evaluate the response behavior and internal mechanism changes of large language models (LLMs) under different prompt settings in medical tasks, we selected the Qwen series of LLMs as the backbone for our experiments, covering various parameter scales. Specifically, we deployed four versions of instruction-tuned models locally: Qwen2.5-7B-Instruct, Qwen2.5-14B-Instruct, Qwen2.5-32B-Instruct, and Qwen2.5-72B-Instruct, representing the performance boundaries of medium, large, and extra-large models within the current open architecture. Additionally, we selected GPT-4o \cite{GPT4} and Deepseek-R1 \cite{deepseek}, two closed-source models based on reasoning-centric architectures that have achieved state-of-the-art (SOTA) results in multiple evaluation lists. These models were used to further explore the impact of different model architectures on role-playing reasoning paths.

The base models used in this study are the Qwen series of LLMs (Qwen2.5-7B/14B/32B/72B-Instruct), developed by Alibaba DAMO Academy. Qwen \cite{qwen}, as one of the leading Chinese open-source models with the highest global download rate, has substantial international influence on the HuggingFace platform and is widely used by both domestic and international research teams in medical question answering systems, virtual doctor assistants, and other scenarios \cite{qwendoctor1, qwendoctor2_qilin, qwendoctor3_medgo}. It has become a core foundational model in the medical LLM ecosystem. The Qwen series is known for its strong multi-turn conversation capabilities, instruction-following ability, and performance in complex contextual understanding and reasoning in medical scenarios \cite{Qwen3}, making it a suitable research platform for role-playing tasks.

GPT-4o \cite{GPT4}, a closed-source multimodal model developed by OpenAI, integrates vision, audio, and text within a unified architecture. Although only its text capabilities were evaluated in this study, its advanced alignment techniques and efficient memory management contribute to its strong instruction-following performance, making it a reliable benchmark for commercial models.

Deepseek-R1 \cite{deepseek}, built on a reasoning-centric architecture with Mixture-of-Experts (MoE), is optimized for retrieval-augmented generation and long-context processing. Its hybrid design reflects the latest trends in scalable LLMs, focusing on high-efficiency reasoning tasks.

To ensure the reproducibility of the experiments and consistency in computational resources, all open-source model evaluations were conducted on high-performance GPU servers of the same type. To enhance the determinism of the experiments and repeatability of the analysis, all model inferences were performed with the random sampling mechanism turned off (do\_sample=False) and using greedy decoding. This setup ensures that the same input generates stable and comparable outputs under different experimental conditions, providing a more accurate reflection of internal representation and behavioral differences under various prompts or neural interventions. For models with extremely large parameters (GPT-4o and Deepseek-R1), we utilized API calls to extract the models' answers to questions and analyzed their performance.

The model invocation interface was built based on HuggingFace Transformers and a local CUDA acceleration framework, and all experiments were completed in the Ubuntu Linux environment to ensure controlled inference response times and efficient resource scheduling. By maintaining a unified model configuration, we further ensured that comparisons across prompts, roles, and experimental tasks are scientifically rigorous and controllable.

\subsection{Neuron Ablation Methods}
To explore the regulatory influence of different role-playing prompts on the model’s internal representation pathways, we designed a controlled ablation experiment framework based on neuron selection and masking. This framework aims to quantify whether role prompts activate specific neurons critical to the decision-making process. The experiment consists of three key components: \textbf{Role-Specific Neuron Selection}, \textbf{Role-Specific Neuron Ablation}, and \textbf{Baseline Ablation Methods}.

\paragraph{Role-Specific Neuron Selection}
% 屏蔽方法图
\begin{figure}
    \centering
    \includegraphics[width=1\linewidth]{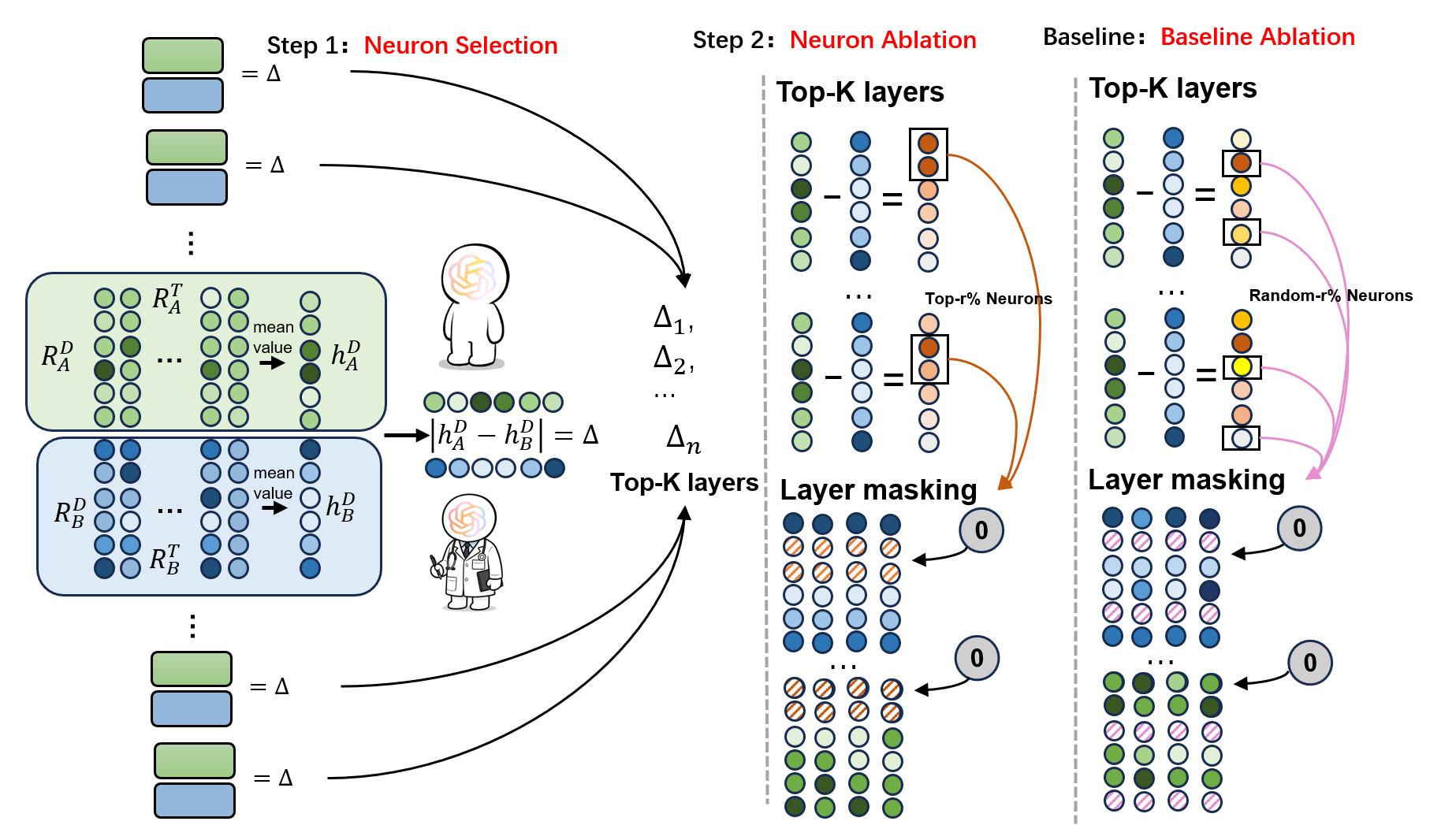}
    \caption{The methods of neurons selection and ablation. It shows the Character Neuron layer selection Method (Step 1), Neuron ablation Method (Step 2) and Baseline Ablation Method(baseline method).}
    \label{fig:ablation_method}
\end{figure}
To identify neurons activated by role-playing prompts, we developed an unsupervised neuron selection method that measures cross-condition activation differences. This method quantifies the impact of role prompts on the model’s internal representations by comparing the hidden state activations under role-playing conditions (with the role prompt) and neutral conditions (without the role prompt).

Let the hidden state outputs under the role-playing condition be denoted as $\mathbf{H}^{\mathrm{role}} = \{ \mathbf{h}^{\mathrm{role}}_l \}_{l=1}^{L}$ and under the baseline condition as $\mathbf{H}^{\mathrm{base}} = \{ \mathbf{h}^{\mathrm{base}}_l \}_{l=1}^{L}$, where $l$ indexes the layers of the model, and $\mathbf{h}_l \in \mathbb{R}^{1 \times T \times d}$ represents the hidden state of the $l$-th layer with token sequence length $T$ and dimensionality $d$. The absolute activation difference at each layer after averaging over tokens is calculated as:
\begin{equation}
\Delta_l = \left| \mathrm{mean}_T(\mathbf{h}^{\mathrm{role}}_l) - \mathrm{mean}_T(\mathbf{h}^{\mathrm{base}}_l) \right| \in \mathbb{R}^d
\end{equation}

Next, we define the role sensitivity score for each layer $l$ as:
\begin{equation}
s_l = \frac{1}{d} \sum_{i=1}^d \Delta_l[i]
\end{equation}

We then rank the layers based on $s_l$ and select the top $K$ layers (with $K$ defaulting to 4). Within each selected layer, we further identify the top $r\%$ (e.g., 5\%) of neurons, forming the cross-layer role-sensitive neuron set $\mathcal{N}_{\mathrm{role}}$. This method allows us to precisely identify the role-specific neurons activated by the role-playing prompts, setting the stage for subsequent neuron ablation.

\paragraph{Role-Specific Neuron Ablation}
To investigate whether the selected role-sensitive neurons are essential for the model's reasoning process, we perform neuron ablation by setting the output activations of these neurons to zero. This simulates the "functional removal" of these neurons and allows us to observe the causal impact on the model's behavior.

Formally, let $\mathcal{N}_l$ represent the set of role-specific neurons identified in the $l$-th layer. We modify the activation tensor $\mathbf{h}^{(l)} \in \mathbb{R}^{T \times d}$ (where $T$ is the token sequence length and $d$ is the hidden space dimensionality) by setting the activations of neurons in $\mathcal{N}_l$ to zero, as follows:

\begin{equation}
\tilde{h}^{(l)}_{t,i} =
\begin{cases}
0, & i \in \mathcal{N}_l \\
h^{(l)}_{t,i}, & \text{otherwise}
\end{cases}
\quad \forall t \in [1, T]
\end{equation}

This ablation method is applied dynamically at each target layer of the Transformer block. It has two main advantages: first, it allows precise identification and manipulation of neural paths without interfering with the overall context distribution; second, it is highly adaptable and cross-model compatible, making it applicable to different architectures and model sizes.

In the experiments, we used a fixed masking ratio (e.g., 1\%) to ensure consistent intervention intensity across different roles. We then compared the accuracy difference between the original and masked models. A significant drop in accuracy indicates that the ablated neurons play a crucial role in the reasoning process under specific roles, revealing their “role-scheduling causality.”

Additionally, in the cross-role transfer experiment, we tested the "cross-masking" setting, where the neurons sensitive to one role are used to mask the reasoning paths of another role. This comparison helps evaluate the generality and selectivity of these neural pathways across different roles.

\paragraph{Baseline Ablation Methods}
To validate the effectiveness of the neuron selection mechanism, we designed baseline ablation schemes. In these schemes, we randomly selected the same proportion of neurons within the same layer range for masking, ensuring that the observed accuracy drop is not solely due to the number of neurons masked, but rather to the specific role-sensitive representation dimensions. By comparing the performance of role-specific ablation with the baseline method, a significantly larger accuracy drop in the former would further support the hypothesis that the selected neurons function as a "behavioral scheduling center" in specific roles.

\subsection{Evaluation Methods}
To comprehensively and rigorously assess the impact of role-playing on the medical reasoning ability of large language models, we designed various quantitative metrics for both behavioral and internal representation layers, supplemented by statistical significance tests and confidence interval analyses to ensure the reliability and repeatability of the results. All open source experiments were conducted on three models with different scales (Qwen2.5-7B-Instruct, Qwen2.5-14B-Instruct, Qwen2.5-32B-Instruct).

\paragraph{Accuracy}
The core evaluation metric for the model's behavior is the answer accuracy. We calculate whether the model's selection tendency in generating answers across different datasets under each role prompt aligns with the correct answer option, defined as:
\begin{equation}
\text{Accuracy} = \frac{\text{Number of Correct Predictions}}{\text{Total Number of Questions}}
\end{equation}

By comparing the accuracy with role prompts to the baseline group (no prompt), we extract the options from the model's output using regular expressions and observe whether role-playing leads to performance improvement or interference.

\paragraph{Jensen-Shannon Divergence}
To measure the extent of changes in the model’s internal representations before and after role-playing, we use JSD\cite{JSD} to quantify the difference in the hidden state vectors in terms of probability distributions. JSD is a symmetric, smoothed variant of the Kullback-Leibler (KL) divergence, defined as:
\begin{equation}
\mathrm{JSD}(P \| Q) = \frac{1}{2} D_{\mathrm{KL}}(P \| M) + \frac{1}{2} D_{\mathrm{KL}}(Q \| M), \quad M = \frac{1}{2}(P + Q)
\end{equation}

Where $P$ and $Q$ represent the hidden state vectors generated by the model under "no role prompt (baseline)" and "with role prompt" conditions, normalized to probability distributions; $M$ is their average distribution; and $D_{\mathrm{KL}}$ is the Kullback-Leibler divergence defined as:
\begin{equation}
D_{\mathrm{KL}}(P \| Q) = \sum_i P(i) \log \frac{P(i)}{Q(i)}
\end{equation}

In the specific implementation, the model generates two versions of hidden layer outputs for the same question (i.e., with a standard prompt vs. with a specific role prompt). We first perform mean pooling for the hidden vectors of each layer, then normalize them to form pseudo-probability distributions. We then calculate the JSD value of the hidden state distribution of each layer under the two conditions, as a quantitative measure of the representation change in that layer.

We calculate the JSD for each sample and role prompt with respect to the baseline for each layer, and then average the JSD results for all samples, plotting a hierarchical curve to analyze the impact range and intensity of different roles on the representation space. This metric is used not only to visually assess whether the model has significantly adjusted its structure for different roles, but also to provide a basis for sensitive layer selection in subsequent neural activation pruning experiments (e.g., role-sensitive neuron ablation).

\paragraph{Principal Component Analysis (PCA)}
To visualize the differences in the model’s hidden states under different role prompts, we perform Principal Component Analysis (PCA) on the hidden states of each role. Let $X \in \mathbb{R}^{n \times d}$ be the matrix of hidden state vectors under the role prompts, and PCA is used to obtain the first two principal components through Singular Value Decomposition (SVD), embedding them in a 2D space to construct a scatter plot. This plot is used to analyze whether different roles exhibit distinct representation clusters or structural boundaries.

\paragraph{Silhouette Coefficient}
After performing K-means dimensionality reduction, we further calculate the silhouette coefficient for each role to quantify the clarity of role clustering. For any data point $i$ of a role, its silhouette coefficient is defined as:
\begin{equation}
s(i) = \frac{b(i) - a(i)}{\max \{a(i), b(i)\}}
\end{equation}
Where $a(i)$ is the average distance from $i$ to other points in its own cluster, and $b(i)$ is the average distance from $i$ to the nearest other cluster. $s(i) \in [-1, 1]$, with a higher value indicating clearer clustering. We use K-means to cluster the roles and report the average silhouette coefficient for all points as a measure of overall semantic distinguishability.

\paragraph{Centered Kernel Alignment (CKA)}
To measure the structural similarity of hidden states, we introduce the CKA analysis method. CKA is a robust neural representation similarity evaluation tool\cite{CKA}, defined as:
\begin{equation}
\mathrm{CKA}(K, L) = \frac{\mathrm{HSIC}(K, L)}{\sqrt{\mathrm{HSIC}(K, K) \cdot \mathrm{HSIC}(L, L)}}
\end{equation}

Where $K$ and $L$ are the Gram matrices of two representation matrices, and HSIC is the Hilbert-Schmidt Independence Criterion kernel norm. We calculate the CKA values for the hidden states of different roles and construct heatmaps to display the degree of role isomorphism in the representation space.

\section*{Results}

\subsection{\textbf{Result 1: Role-Playing Does Not Enhance the Medical Reasoning Capabilities of Large Language Models}}
\begin{figure}
    \centering
    \includegraphics[width=0.6\linewidth]{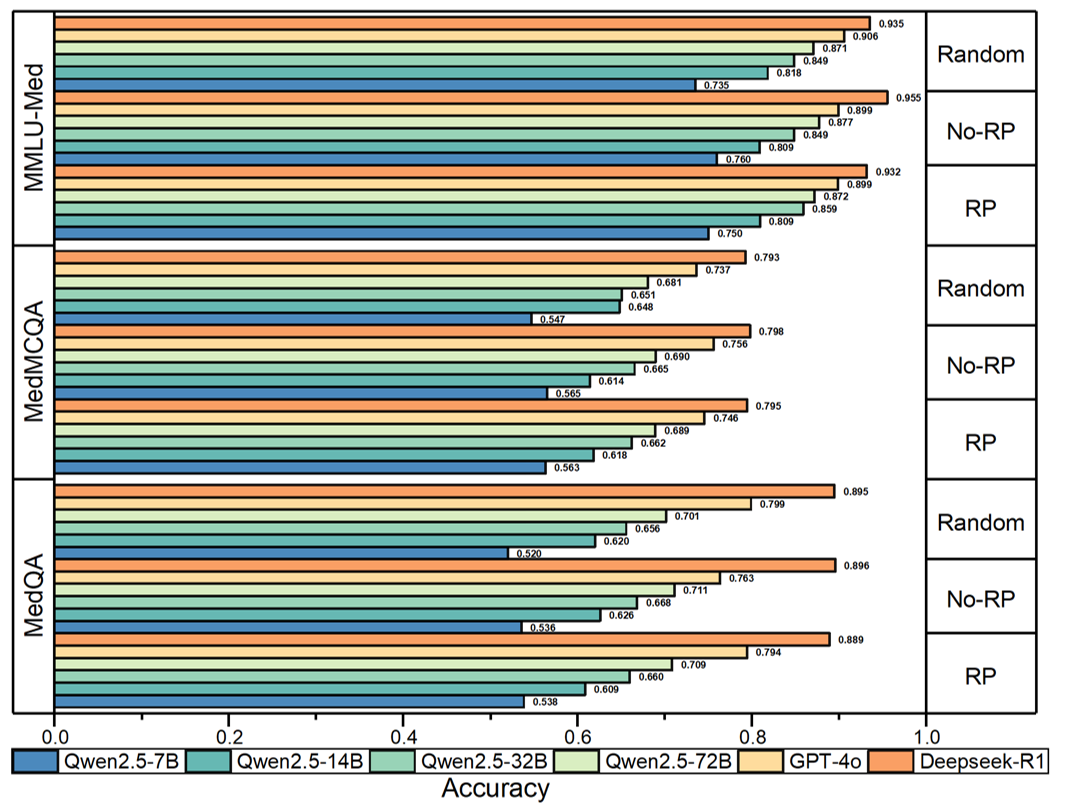}
    \caption{QA accuracy of 6 kinds of LLM on 3 kinds of medical QA datasets}
    \label{fig:average_acc}
\end{figure}

We began by validating a core hypothesis: if role-playing prompts can truly enhance the medical reasoning capabilities of LLMs, the model's answer accuracy should significantly improve under such guidance. To investigate this, we designed a systematic comparative experiment to analyze the QA accuracy of the test.

We tested the answer accuracy of different model scales (Qwen2.5-7B/14B/32B/72B-Instruct, GPT-4o, and Deepseek-R1) on the same set of medical multiple-choice questions, under three types of Prompt groups:
1. Role-Play Group (RP): 7 different doctor identity descriptions for role-playing.
2. Baseline Group (No-RP): Direct model guidance without role-playing prompts.
3. Random Group: Randomly generated statements replacing the role-playing prompts.

\begin{figure}
    \centering
    \includegraphics[width=1\linewidth]{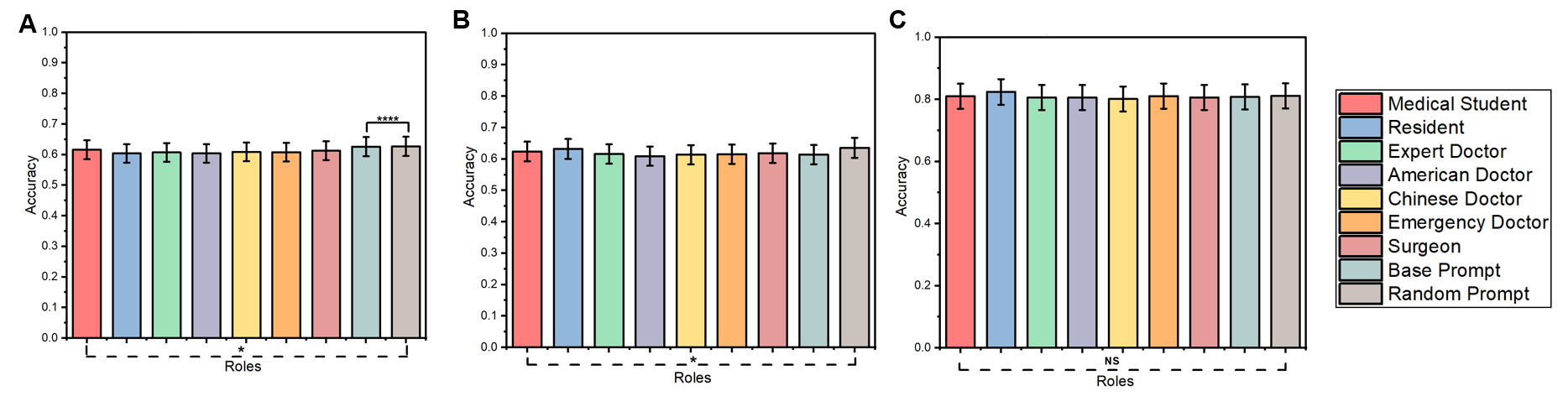}
    \caption{QA accuracy of 9 roles play results tested on 3 kinds of medical datasets based on Qwen2.5-14B-Instruct}
    \label{fig:role_acc2}
\end{figure}

As shown in Figure \ref{fig:role_acc2}. The accuracy evaluation using 7 different doctor role-playing prompts on 6 types of LLMs demonstrates that role-playing (RP) did not significantly improve the model's answer accuracy. In the MedMCQA and MedQA datasets, accuracy differences between RP and Random or No-RP were negligible, with a variation of less than ±2\%. In the MMLU-Med dataset, the performance of the doctor role-playing group showed almost no difference, indicating that adopting a clinician's perspective did not significantly affect the model's medical reasoning capabilities.We used Cochran's Q test to further validate the accuracy differences between the RP group and the baseline group (No-RP). The results showed that in most cases (p > 0.05), the differences were not statistically significant, indicating that role prompts did not significantly enhance the model’s medical reasoning capabilities."

To further explore whether specific doctor roles could enhance reasoning effects, we display the accuracy distributions of 3 open-source models (Qwen2.5-7B/14B/32B-Instruct) under the 7 doctor role prompts (Figure \ref{fig:role_acc2}). Significance test analysis, using Cochran’s Q test (indicated by dashed lines), revealed no significant differences between the majority of the role groups (p > 0.05). For those groups with significant Cochran’s Q test results, pairwise McNemar tests (after Holm correction) showed no statistical significance in most cases (p > 0.05).

Furthermore, when examining the impact of model architecture and reasoning methods on role-playing, we observed no significant accuracy differences between models, regardless of whether the model was role-playing as a "medical student" or an "expert doctor" (Table \ref{reasoning models}). This consistent trend across different tasks suggests that large language models do not develop distinct reasoning response patterns based on different role settings.

These results conclude that role-playing prompts do not activate or enhance the model’s medical reasoning capabilities. Instead, they appear to primarily alter linguistic style, rather than influencing the internal decision-making process of the model.
\begin{table}
\centering
\caption{QA performance comparison of 3 kinds of models with different architecture on MedQA dataset}
\begin{tabular}{l*{8}{c}}
\toprule

\makecell{\textbf{Model}\\\textbf{Name}} & 
\makecell{\textbf{Model}\\\textbf{Architecture}} & 
\makecell{\textbf{Medical}\\\textbf{Student}} & 
\makecell{\textbf{Resident}} & 
\makecell{\textbf{Expert}\\\textbf{Doctor}} & 
\makecell{\textbf{American}\\\textbf{Doctor}} & 
\makecell{\textbf{China}\\\textbf{Doctor}} & 
\makecell{\textbf{Emergency}\\\textbf{Doctor}} & 
\makecell{\textbf{Surgeon}} \\
\midrule
Qwen2.5-72B&Reasoning base(Dense) & 0.7227 & 0.7054 & 0.7069 & 0.714 & 0.7046 & 0.7046 & 0.7014 \\
GPT-4o&Closed Source & 0.802 & 0.7824 & 0.7824 & 0.8114 & 0.8067 & 0.7824 & 0.7941 \\
Deepseek-R1&Reasoning base(MoE) &\textbf{ 0.8939} & \textbf{0.8900} & \textbf{0.8982} & \textbf{0.8955} & \textbf{0.8884} & \textbf{0.8892} & \textbf{0.8884} \\

\bottomrule
\end{tabular}
\label{reasoning models}
\end{table}

\subsection{\textbf{Result 2: Role-Playing Doctors Follow Highly Similar Cognitive Pathways}} \label{neurons}

In clinical practice, doctors at different levels (e.g., interns, residents, attending physicians) exhibit distinct cognitive pathways and task strategies. However, current LLMs show highly consistent reasoning pathways, regardless of the doctor role they are simulating.

To investigate whether LLMs exhibit role-specific reasoning pathways, we designed a neuron masking intervention. We first conducted a parameter sensitivity analysis to validate the general effectiveness of this intervention, with results detailed in Table \ref{tab:ablation-strength}. This analysis confirms a clear dose-response relationship: increasing the ablation strength---either by targeting more layers (from Top-4 to Top-8) or increasing the masking percentage (from 3\% to 10\%)---consistently results in a more significant accuracy drop across all three datasets. For instance, on the MedQA dataset, the accuracy drop for Top-8 layers at 10\% (0.547) is notably larger than that for Top-4 layers at 3\% (0.457).

Having established the intervention's efficacy, we selected a moderate and representative parameter set for our primary role-specific experiments: masking the \textbf{top 5\%} most significantly activated neurons from the \textbf{top 4 hidden layers}. We tested this experiment on three different model sizes (Qwen2.5-7B-Instruct, Qwen2.5-14B-Instruct, and Qwen2.5-32B-Instruct).

\begin{figure}
    \centering
    \includegraphics[width=1\linewidth]{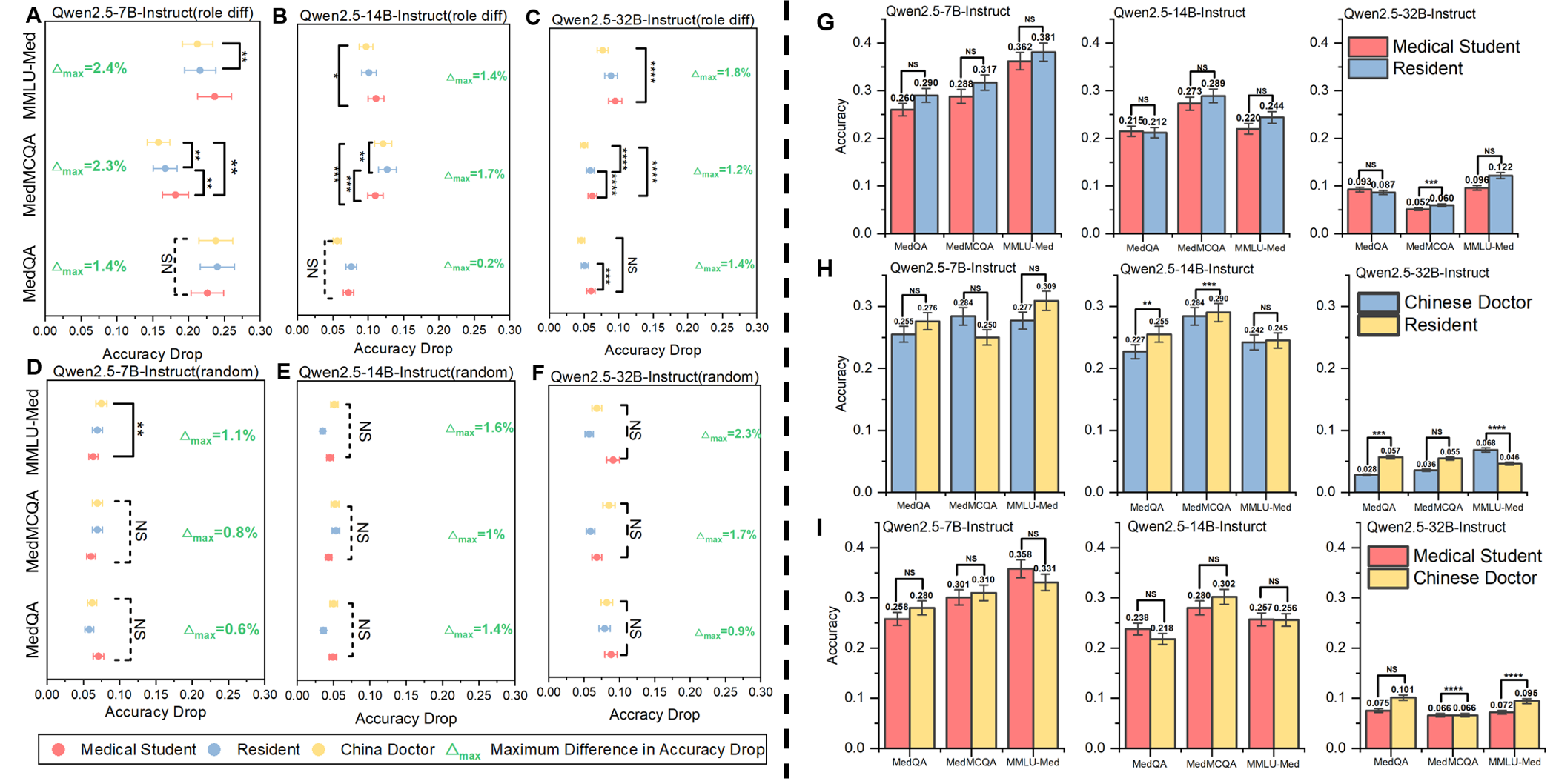}
    \caption{Accuracy drop between RP LLM and No-RP LLM(A-F), figure G-I shows the accuracy between cross ablation experiments."role\_diff" means the accuracy drop by RP based ablation method, "random" means the accuracy drop by random ablation method}
    \label{fig:RoleAccDrop}
\end{figure}

As shown in Figures \ref{fig:RoleAccDrop}.A-F , the accuracy drop after masking was consistent across all roles. For example, in the Qwen2.5-32B-Instruct model on the MedQA dataset, the accuracy drops for the roles of "Resident," "Medical Student," and "China Doctor" were 0.051, 0.060, and 0.046, respectively. The maximum difference in accuracy drop was only 1.4\%, and the same pattern was observed across different datasets (MedQA, MedMCQA, MMLU-Med). We also conducted McNemar's test and observed no significant differences in accuracy drops between roles ($p > 0.05$). This indicates that the impact of neuron ablation on performance was consistent across different roles.

Further cross-role neuron masking experiments (Figures \ref{fig:RoleAccDrop}.G-I) showed that the performance drops between roles were almost identical, with no significant differences in accuracy. For example, the cross-masking of "Medical Student" and "Resident" roles led to negligible performance differences (Qwen2.5-14B: $p = 0.317$; 32B: $p = 0.362$, McNemar test).

These results suggest that LLMs do not construct differentiated cognitive pathways when simulating different doctor roles. Instead, they rely on similar underlying activation structures, regardless of the role. This high overlap in cognitive pathways suggests that current LLMs cannot simulate distinct thinking patterns for different roles, and their responses are primarily controlled by linguistic style rather than cognitive differentiation.

\begin{table}[t]
  \centering
  \setlength{\tabcolsep}{8pt}        % 列间距
  \renewcommand{\arraystretch}{1.15}  % 行距
  \begin{tabular}{lcccc}
    \toprule
    \textbf{Ablation Layers} & \textbf{Ablation Percentage per layer} & \textbf{MedQA} & \textbf{MedMCQA} & \textbf{MMLU-Med} \\
    \midrule
    \multirow{3}{*}{Top-4 layers}
      & 3\%  & 0.457 & 0.496 & 0.624 \\
      & 5\%  & 0.460 & 0.502 & 0.653 \\
      & 10\% & \textbf{0.493} & \textbf{0.543} & \textbf{0.704} \\
    \midrule
    \multirow{3}{*}{Top-6 layers}
      & 3\%  & 0.427 & 0.489 & 0.599 \\
      & 5\%  & 0.436 & 0.493 & 0.626 \\
      & 10\% & \textbf{0.518} & \textbf{0.559} & \textbf{0.731} \\
    \midrule
    \multirow{3}{*}{Top-8 layers}
      & 3\%  & 0.512 & 0.546 & 0.703 \\
      & 5\%  & 0.528 & 0.563 & 0.744 \\
      & 10\% & \textbf{0.547} & \textbf{0.574} & \textbf{0.757} \\
    \bottomrule
  \end{tabular}
  \caption{Accuracy drop at different ablation strengths and per-layer percentages on MedQA, MedMCQA, and MMLU-Med.}
  \label{tab:ablation-strength}
\end{table}

\subsection{\textbf{Result 3: Role-Playing Doctors Lack the Ability to Model Hierarchical Structures in Medical Professional Levels}} \label{layers}
\begin{figure}
    \centering
    \includegraphics[width=1\linewidth]{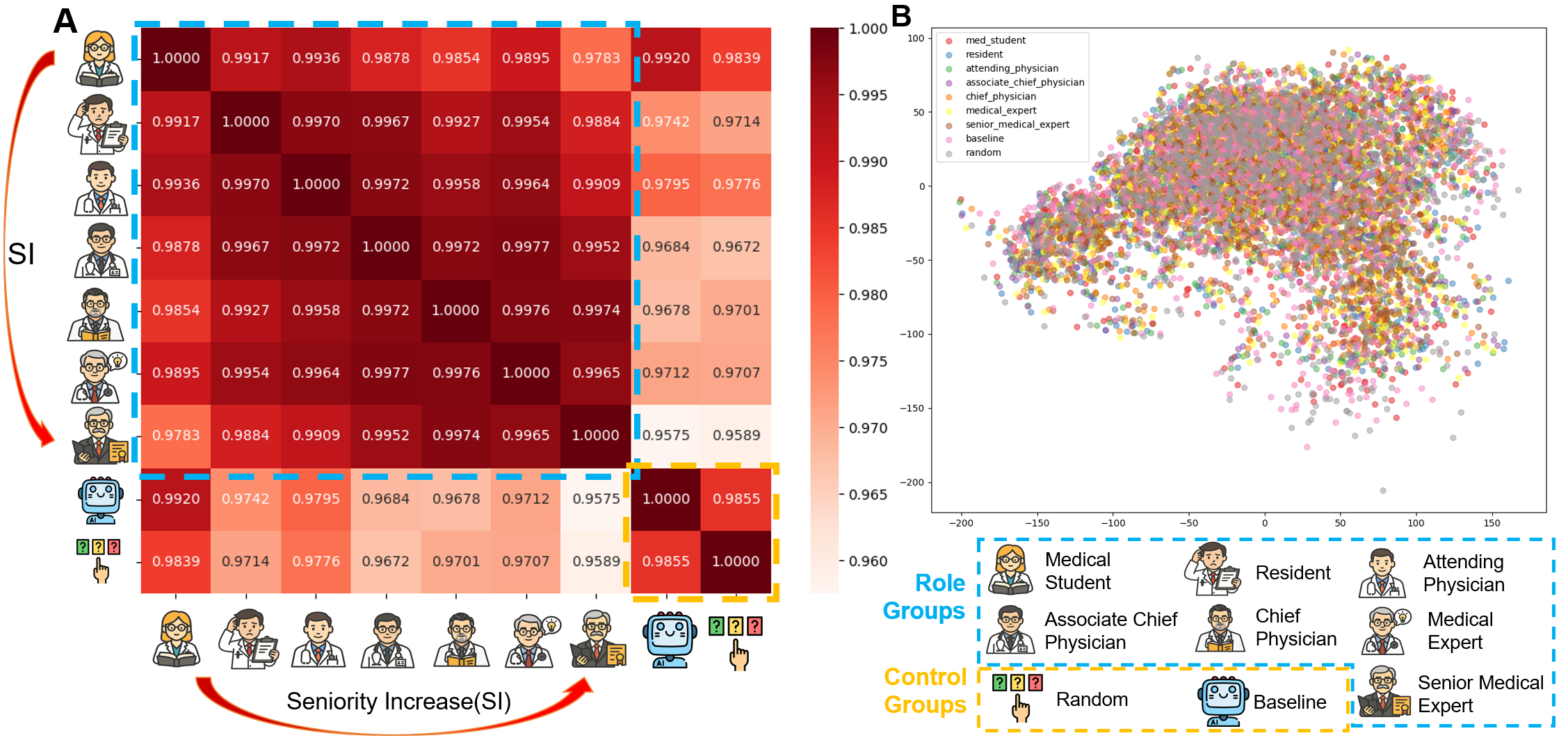}
    \caption{The CKA analysis and PCA analysis of role play in seniority increase scanario. Subfigure A shows the CKA similarity heatmap of activations between different doctor roles, displaying the average hidden layer similarity of Qwen2.5-14B-Instruct model under different medical role prompt inputs. Subfigure B shows the PCA visualization of the final hidden layer output for different roles in MedQA dataset.}
    \label{fig:pca_ck}
\end{figure}
\begin{table}[htbp]
\centering
\caption{Silhouette scores for different roles in Qwen2.5-14B model in MedQA benchmark.}
\begin{tabular}{l c}
\toprule
\textbf{Role} & \textbf{Silhouette Score} \\
\midrule
Medical Student & 0.12 \\
Resident & 0.13 \\
Attending Physician & 0.13 \\
Chief Physician & 0.13 \\
Associate Chief Physician & 0.13 \\
Medical Expert & 0.13 \\
Senior Medical Expert & 0.14 \\
Baseline & 0.11 \\
Random & 0.11 \\
\bottomrule
\end{tabular}
\label{SC}
\end{table}

In clinical practice, doctors at different levels of expertise use different cognitive strategies based on their professional ranking. However, we found that LLMs, when given role-playing prompts for different doctor levels, fail to form distinguishable professional cognitive structures. Their activation patterns remain highly similar across roles.

We used the CKA metric to measure the similarity of hidden layer representations induced by different doctor role prompts. As shown in Figure \ref{fig:pca_ck}.A, the similarity between different roles was consistently high (0.96–1.00), indicating that the representation space for different roles overlaps significantly. There was no clear gradient in activation structures along the professional hierarchy.

Further, Principal Component Analysis (PCA) of the final hidden layer representations (Figure \ref{fig:pca_ck}.B) showed that the samples for different roles heavily overlapped in the projection space, exhibiting almost no visible clustering boundaries. The role-specific clustering results based on K-means, with similar silhouette scores for all roles(as shown in Table.\ref{SC}), confirmed that the model failed to differentiate between roles based on the professional hierarchy.

This analysis indicates that LLMs do not generate distinct cognitive structures for different medical roles. Instead, the role-playing prompts only influence surface-level language features, failing to trigger deeper structural differentiation. This highlights a significant limitation in current LLM-based role-playing: while the models can mimic the linguistic style of different roles, they do not reproduce the hierarchical cognitive differentiation expected of human doctors.

\subsection{\textbf{Result 4: Role-Playing Doctors Do Not Simulate Specific Cognitive Pathways of Real Doctors}} \label{JSD}

To evaluate whether role-playing prompts trigger structural changes in the reasoning pathways of LLMs, we used JSD as an indicator to systematically quantify the deviation of hidden state representations in Transformer layers across different role conditions. We conducted this experiment on the MedMCQA dataset, using three sizes of Qwen2.5 series models (7B-Instruct, 14B-Instruct, 32B-Instruct).

\begin{figure}
    \centering
    \includegraphics[width=1\linewidth]{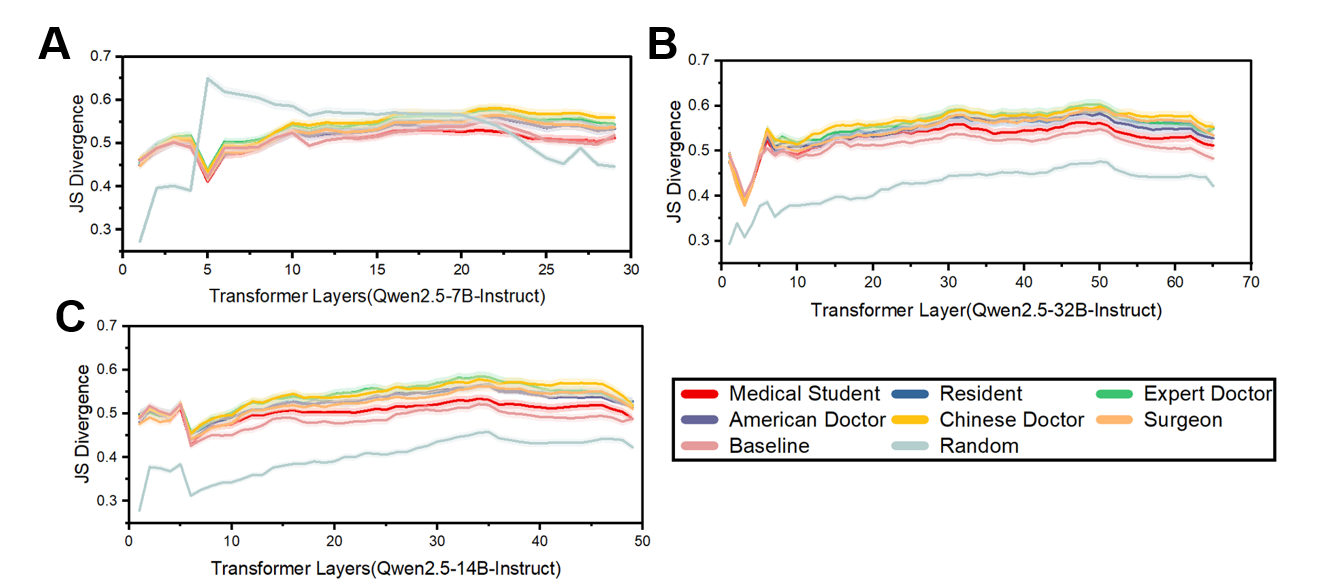}
    \caption{Experiment results of visualizing the JS divergence of different kinds of RP LLMs,figure A-C show the JSD between different layers}
    \label{fig:JS layers2}
\end{figure}

Figures \ref{fig:JS layers2}.A–C show that the significant differences in JSD between the RP group and the control groups (Baseline and Random) were concentrated in the model's shallow-to-mid layers (e.g., layers 1–25 in Figure \ref{fig:JS layers2}.C). This suggests that role prompts affect the early language modeling and semantic encoding processes of the model. However, as the Transformer layers deepen, the representation differences diminish and converge (after layer 35 in Figure \ref{fig:JS layers2}.C). This indicates that role prompts do not have lasting effects on the deeper layers of the model.

This suggests that while role settings can influence the model's initial reasoning or language expression, they do not significantly alter the core reasoning pathways. This pattern holds for all roles, indicating that LLMs do not exhibit distinct cognitive strategies when playing different doctor roles, and their reasoning paths remain largely similar.

These results demonstrate that LLMs do not replicate real-world human doctors' cognitive processes but merely simulate language styles. It highlights the limitations of using role-playing prompts to induce real cognitive differentiation in medical AI.

\section*{Discussion}
This study systematically evaluated the impact of role-playing prompts on the reasoning capabilities of LLMs in medical question answering tasks. Our results reveal that while role-playing prompts significantly alter the language style of the models, they do not induce meaningful differences in the cognitive reasoning pathways expected from different medical professionals. Specifically, despite role-playing prompts simulating various doctor roles, such as medical student, resident, and attending physician, the model's underlying decision-making processes remained strikingly similar across all roles. This finding challenges the common assumption that "language equals cognition," particularly in the context of medical AI, where the cognitive differences between medical professionals are crucial for accurate diagnosis and treatment.

A central question driving this research was whether role-playing could induce role-specific reasoning pathways in LLMs, analogous to the cognitive strategies used by human doctors of different expertise levels. However, our findings indicate that role-playing primarily impacts the model's surface-level language output, without triggering the activation of distinct reasoning circuits. This is evident from our experiments with neuron ablation, where blocking role-specific neurons did not lead to significant differences in the model's performance. The minimal performance drop after neuron masking across different roles suggests that the model does not develop role-specific cognitive pathways, but instead relies on a monolithic reasoning mechanism that operates similarly across different professional roles.

Moreover, our analysis of the model’s hidden state representations at different layers revealed no evidence of hierarchical differentiation between doctor roles. While role prompts did influence the early layers of the Transformer, these effects dissipated in the deeper layers, where the model’s representations converged. This suggests that although role prompts can influence the model's initial response, they do not lead to substantial changes in the core reasoning structure, which remains largely unaffected by the professional role of the simulated doctor.

One of the key takeaways from this study is that current prompt-based role-playing methods are insufficient for replicating the nuanced cognitive differences observed in real-world clinical decision-making. In a medical setting, different roles involve not just differences in knowledge but also differences in cognitive strategies, prioritization, and clinical judgment. The lack of cognitive differentiation in LLMs, as shown by our experiments, implies that relying on role-playing to simulate clinical expertise could lead to misleading or unreliable medical decision support. Specifically, in complex clinical scenarios requiring differentiated reasoning, such as diagnosing rare conditions or managing high-risk patients, the model may fall short of human-level reasoning abilities, potentially posing risks to patient safety.

Our findings also underscore the limitations of the current paradigm in clinical AI, which often equates the language produced by a model with the reasoning abilities associated with that language. Although LLMs have demonstrated impressive performance in language generation tasks, their current architectures do not enable them to truly simulate the mental models of medical professionals. Role-playing prompts may shape how the model expresses itself but do not equip it with the ability to reason like a medical expert. This distinction is critical, as it highlights the gap between superficial language imitation and genuine cognitive modeling.

Looking ahead, the future of medical AI should focus on developing models that go beyond language imitation and incorporate cognitive modeling techniques. Rather than simply simulating roles, future models should aim to understand and replicate the cognitive processes underlying clinical decision-making. This could involve integrating domain-specific knowledge into the training process and designing systems that can simulate the reasoning strategies of different medical professionals. Such advancements would pave the way for more reliable and interpretable AI systems capable of providing professional-level medical decision support.

While role-playing in LLMs can alter language style, it does not provide the model with the deeper cognitive abilities required for accurate medical decision-making. Our study calls for a shift in how we approach the development of clinical AI, advocating for models that simulate cognitive processes rather than just linguistic behavior. By doing so, we can better align AI systems with the complexities of real-world clinical practice, ensuring their utility and safety in medical applications.
\section*{Conclusion}
This study systematically evaluated the impact of role-playing prompts on the reasoning capabilities of large language models in medical question answering tasks. The results reveal that, while role-playing prompts can significantly alter the model's language style, they do not induce reasoning pathways that correspond to the professional roles of doctors. Despite testing across multiple datasets, role-playing failed to improve the model’s accuracy, particularly in high-complexity tasks. Instead, the effect of role-playing was limited to superficial language changes, with no activation of new knowledge structures or reasoning strategies. Furthermore, neuron ablation and hierarchical modeling tests demonstrated that the reasoning pathways across different doctor roles were nearly identical, indicating a lack of role-specific cognitive differentiation. These findings highlight the limitations of current role-playing methods in clinical AI, as they primarily alter language behavior rather than enabling models to replicate the cognitive processes of real doctors. Consequently, relying on language-based role-playing alone is insufficient for building reliable cognitive agents in medical applications, signaling the need for a shift towards more advanced cognitive modeling in the development of medical AI systems.
\bibliography{reference}
\section{Data availability}
The datasets used in this study are publicly available. The MedQA dataset (USMLE subset) can be accessed at \\ \url{https://github.com/jind11/MedQA} and includes US medical licensing examination questions for clinical reasoning evaluation. The MedMCQA dataset is sourced from Indian medical exams and is available at \url{https://github.com/medmcqa/medmcqa}. The MMLU-Med subset, comprising interdisciplinary medical knowledge tasks, can be found at \url{https://github.com/hendrycks/test}.

All code for reproducing our analysis is available in the following repository:\url{https://github.com/IAAR-Shanghai/RolePlay_LLMDoctor#}

\section*{Inclusion \& Ethics Statement}
This study was conducted with a strong commitment to ethical research practices and inclusive scientific inquiry. Our research focuses on the cognitive modeling capabilities of large language models (LLMs) in the context of medical role simulation. No human subjects or private health data were used or involved at any stage of the study. All datasets employed—MedQA, MedMCQA, and MMLU-Med—are publicly available and accessed under their respective open licenses, ensuring transparency and reproducibility.

The authors affirm that no discriminatory, exclusionary, or harmful content is present in the datasets, prompts, or experimental design. The roles designed in this study span diverse medical backgrounds and geographies (e.g., doctors from China, the U.S., and various medical ranks), reflecting our intent to ensure inclusivity in professional representation.
\section*{Acknowledgements}
This work is supported by the National Natural Science Foundation of China under grant 62072463, the Scientific Research Fund of Renmin University of China (Central Universities Basic Scientific Research Funds) under project 24XNKJ31, and the Open Fund of the National Key Laboratory of Digital Publishing Technology, Founder Group. The corresponding authors of this paper are Yanfang Chen and Zhiyu Li.
 
\section*{Author contributions statement}
X,L. Z.L. and H.L. conceived the study and supervised the overall project. H.L. and H.W. designed and executed the core experiments, including the role prompt construction, accuracy evaluation, and neural representation analyses. W.Z. and L.Z. implemented the neuron ablation framework and conducted the role-specific masking experiments. Y.C. curated and processed the medical QA datasets and contributed to the design of the cognitive stratification assessment. F.X., H.L., X.L., H.W., and W.Z. jointly analyzed the results and drafted the manuscript. Z.L. and Y.C. provided critical feedback on the experimental methodology and clinical implications. All authors contributed to the revision of the manuscript and approved the final version.

\section*{Competing interests}  
The authors declare no competing interests.

\end{document}